%% file: iclr2026_conference.tex
\documentclass{article} %
\usepackage{iclr2026_conference,times}
\iclrfinalcopy

\input{math_commands.tex}

\usepackage[utf8]{inputenc} %
\usepackage[T1]{fontenc}    %
\usepackage{hyperref}       %
\usepackage{url}            %
\usepackage{booktabs}       %
\usepackage{amsfonts}       %
\usepackage{nicefrac}       %
\usepackage{microtype}      %
\usepackage{xcolor}         %

\usepackage{graphicx} %
\usepackage{booktabs,adjustbox}
\usepackage{makecell}

\usepackage{siunitx}        %
\usepackage{amsthm}

\usepackage{listings}

\usepackage{rotating}
\usepackage{multirow}

\definecolor{codegreen}{rgb}{0,0.6,0}
\definecolor{codegray}{rgb}{0.5,0.5,0.5}
\definecolor{codepurple}{rgb}{0.58,0,0.82}
\definecolor{backcolour}{rgb}{0.95,0.95,0.92}

\lstdefinestyle{mystyle}{
    backgroundcolor=\color{backcolour},   
    commentstyle=\color{codegreen},
    keywordstyle=\color{magenta},
    numberstyle=\tiny\color{codegray},
    stringstyle=\color{codepurple},
    basicstyle=\ttfamily\footnotesize,
    breakatwhitespace=false,         
    breaklines=true,                 
    captionpos=b,                    
    keepspaces=true,                                     
    numbersep=5pt,                  
    showspaces=false,                
    showstringspaces=false,
    showtabs=false,                  
    tabsize=2
  }

\title{WRAVAL – WRiting Assist eVALuation}

\author{%
  Gabriel Benedict\thanks{In order of joining the project.} \\
  Amazon Fire Tablets \\
  Madrid, Spain \\
  gbndict@amazon.es \\
  \And
  Matthew Butler \\
  Amazon Devices Product \\
  Nova Scotia, Canada \\
  matbutle@amazon.com \\
  \And
  Naved Merchant \\
  Amazon Devices OS \\
  Sunnyvale, USA \\
  mernaved@amazon.com \\
  \And  
  Eetu Salama-Laine \\
  Amazon Fire Tablets \\
  Madrid, Spain \\
  eetsala@amazon.es \\
}

\begin{document}
\fancyhf{}  %
\renewcommand{\headrulewidth}{0pt}  %

\maketitle

\begin{abstract}
The emergence of Large Language Models (LLMs) has shifted language model evaluation toward reasoning and problem-solving tasks as measures of general intelligence. Small Language Models (SLMs) – defined here as models under 10B parameters – typically score 3-4 times lower than LLMs on these metrics. However, we demonstrate that these evaluations fail to capture SLMs' effectiveness in common industrial applications, such as tone modification tasks (e.g., funny, serious, professional). We propose an evaluation framework specifically designed to highlight SLMs' capabilities in non-reasoning tasks where predefined evaluation datasets don't exist. Our framework combines novel approaches in data generation, prompt-tuning, and LLM-based evaluation to demonstrate the potential of task-specific finetuning. This work provides practitioners with tools to effectively benchmark both SLMs and LLMs for practical applications, particularly in edge and private computing scenarios. Our implementation is available at: \url{https://github.com/amazon-science/wraval}.
\end{abstract}

\section{Introduction}

Writing assistance tools powered by Language Models (LMs) are becoming increasingly prevalent on personal devices~\cite{gemini, appleIntelligence, phiSilica}. These applications typically use Small Language Models (SLMs) with fewer than 8B parameters due to memory and latency constraints of mobile devices. While SLMs may not match Large Language Models (LLMs) in complex reasoning tasks\cite{SLMReasoningFT, SLMReasoning, phi4MiniReasoning}, they are finding widespread adoption in focused applications like text rewriting.

Despite this growing deployment of SLMs, existing benchmarks predominantly focus on evaluating LMs' reasoning capabilities (e.g., MMLU~\cite{mmlu}, SuperGLUE~\cite{superGlue})\footnote{In addition to Math and STEM tasks (e.g. GPQA~\cite{gpqa}), and coding tasks (e.g. evalPlus~\cite{evalPlus}).}. This creates a significant gap: we lack standardized ways to evaluate SLMs on many of their typical use cases - including writing assistance tasks. We propose that writing assistance tasks represent a distinct category of language tasks where parameter efficiency can be achieved more readily than in reasoning tasks, as they primarily involve pattern recognition and stylistic transformation rather than complex logical operations across multiple inference steps. This efficiency difference may explain why SLMs can perform competitively on these tasks despite their parameter constraints.

\begin{figure}
    \centering
    \includegraphics[width=\linewidth]{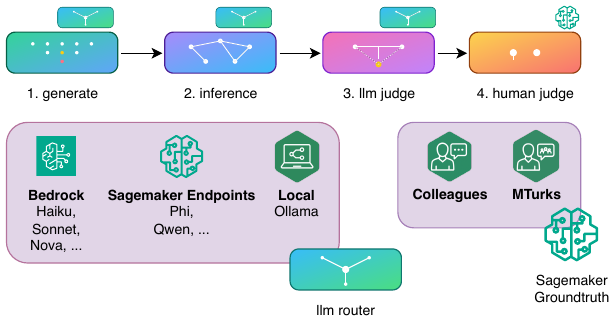}
    \caption{WRAVAL – Writing Assistant eVALuation pipeline. Our proposed framework (1) generates data with custom prompts at scale, (2) performs inference with models hosted on Bedrock / Ollama / Sagemaker Endpoints, (3) runs A bigger LLM-as-a-judge on all inputs-output pairs, (4) transforms the data into a format ready for human judges. A symbol hovers over each step of the pipeline to indicate that it is supported by either our own llm router (bottom left), or Sagemaker Groundtruth (bottom right).}
    \label{fig:wraval-pipeline}
\end{figure}

In this paper, we address this gap by introducing a novel evaluation framework for Writing Assistance (WA) tasks, which we define as single-turn text transformations guided by specific instructions. Our framework focuses on nine common rewrite instructions: casual, elaborate, emojify, improve, keypoints, professional, proofread, shorten and witty (see \url{https://github.com/amazon-science/wraval} for tone definitions). Below is an example professionalized message, and a verbatim output of the command \texttt{wraval show-examples}:

\begin{lstlisting}
=========================================================================
Tone: professional | Model: qwen-3-4B
=========================================================================
Example 7545:
Original: I was feelin' myself in that outfit, bruh, no lie.
Rewrite:  I felt confident in that outfit - no doubt about it.
Score:    3.00
----------------------------------------
\end{lstlisting}

Our framework departs from traditional static benchmarks by dynamically generating both evaluation data and assessments, enabling task-specific personalization. Using this approach, we demonstrate that SLMs are narrowing the performance gap with LLMs on writing assistance tasks. Additionally, our findings reveal unexpected limitations of LLMs in certain non-reasoning scenarios, challenging conventional assumptions about their superiority across all tasks (see Section~\ref{sec:results}).

\section{Related Work}

\paragraph{SLMs on non-reasoning tasks} Small Language Models (SLMs) have demonstrated increasing performance on non-reasoning tasks, maintaining coherence in English text generation and handling simpler tasks such as extraction and summarization effectively. For instance, models like Phi-1 have shown promising results in these capacities without extensive reasoning capabilities \cite{tinystories, phi1}. This suggests a theoretical distinction between reasoning and non-reasoning language tasks, where the latter may require less computational complexity and can be effectively performed with fewer parameters. The parameter efficiency of non-reasoning tasks may be explained by their reliance on pattern recognition and application of learned linguistic conventions rather than multi-step logical inference.

\paragraph{Evaluation Frameworks} Robust evaluation of text improvement has been advanced by benchmarks such as EDITEVAL, which focuses on instruction-based assessments of modular text editing skills, including cohesion improvements and paraphrasing~\cite{editeval}. Additionally, Apple has developed a rigorous evaluation process using highly trained human graders with writing and comprehension expertise to assess summarization and naturalness of generated text across locales and tasks~\cite{appleIntelligence}.

\paragraph{Evaluation Datasets} Publicly available benchmark datasets play a pivotal role in evaluating language model capabilities across various tasks, including translation, summarization, and text improvements. For translation, the annually released Conference on Machine Translation (WMT) dataset provides multiple language pairs; our work adopts the WMT16 dataset specifically for German-to-English translation as an initial benchmark~\cite{wmt16}. In summarization, widely used datasets such as Gigaword and SAMSum enable assessment of models' ability to generate concise and coherent summaries \cite{samsum}. For text improvements, datasets like EDITEVAL offer instruction-based evaluations, while MESSAGEREWRITEEVAL builds upon EDITEVAL by fine-tuning models on five distinct rewriting tasks—formalization, shortening, elaboration, paraphrasing, and proofreading—using task-specific instruction prompts and human judgment guidelines \cite{editeval, messageRewrite}.

\paragraph{Finetuning} Parameter-efficient fine-tuning techniques such as LoRa have been successfully applied for multilingual summarization tasks, enabling effective model adaptation with fewer trainable parameters \cite{loraSummaries}. Reinforcement learning (RL) methods have been utilized for various text rewriting tasks including formalization, shortening, elaboration, paraphrasing, and proofreading. Such RL-driven finetuning has shown improvements in task-specific performance, with dedicated studies further focusing on optimizing proofreading via RL \cite{messageRewrite, proofreadFinetune}.

\paragraph{} Despite these resources, the limited scope of existing frameworks and datasets for non-reasoning tasks motivate the use of generated data and our own framework to further advance model training and evaluation in this domain.

\section{Design}

This section is dedicated to the framework, and its associated experimental design.

\subsection{Framework – WRAVAL: Writing Assistant eVALuation }

WRAVAL is an open-source library designed to evaluate Language Models (LMs) on Writing Assistance (WA) tasks, such as text professionalization and summarization. The framework implements a sequential workflow managed through a centralized table structure, where rows represent new data and columns represent workflow steps. Version control is maintained by time-stamping each table iteration, with flexible storage options for both local and cloud environments. The workflow comprises four primary steps illustrated in Figure~\ref{fig:wraval-pipeline}: data generation (\texttt{generate}), inference processing (\texttt{inference}), LLM-based evaluation (\texttt{llm judge}), and human judgment (\texttt{human judge}). The \texttt{lm router} and \texttt{prompt directory} components complement these core functions. Below, we detail each component before presenting recommended implementation guidelines.

\paragraph{\texttt{generate}}

The generation step is responsible for generating a synthetic dataset with an LM. A single execution of this step can generate data either for a single tone or all possible tones. The prompt is of the form "generate data that could be turned more professional". We provide code to generate around a 100 (default value) examples with one prompt and code to generate a heterogeneous set of examples via multiple prompts (see Section~\ref{sec:learnings}). We ask the model to generate CSV, which we parse into a table. That table serves as our database, as mentioned above.

\paragraph{\texttt{inference}}

The previous step determines what example should be rewritten with what tone (with a tone column). The inference step consumes the synthetic data and their associated potential tone change from the previous step, and rewrites them with an LM. 

\paragraph{\texttt{llm judge}}

The LLM judge step consumes the input from \texttt{generate} and the output from \texttt{inference}, and produces a numerical score describing the quality of the rewrite obtained from the \texttt{inference} step. We specifically recommend a reference LLM here as opposed to an SLM. Via prompt engineering, the user can ensure that judge outputs judgments and justifications on different aspects (e.g. clarity, conciseness), and a score enclosed in brackets. Our code assumes that format to then extract the score. We provide working prompts in the open source code.

\paragraph{\texttt{human judge}}

The human judge step, like for \texttt{llm judge}, takes the results from \texttt{generate} and \texttt{inference}. It then submits them to Sagemaker Groundtruth. The human judge can be first colleagues, then MTurks (both possible via our framework using Sagemaker Groundtruth).

\paragraph{\texttt{lm router}}

For the first three steps, we use an LM router: the user can decide to run an LLM on Bedrock (Claude~\cite{claude}, ChatGPT~\cite{chatGPT}, ...), a quantized SLM on a Sagemaker Endpoint, or a local SLM with Ollama~\cite{ollama}. At each step of the workflow, we record the model used in a column. For each solution, we propose both sequential and batch processors. While Bedrock and SageMaker provide built-in batch processing, we found that for datasets under 100 examples, parallelizing API calls via Python multiprocessing is more efficient than using their native batch processors.

\paragraph{\texttt{prompt directory}}

The proposed WRAVAL framework is prompt agnostic. We manage prompts in a versioned and object-oriented way that guarantees traceability and adaptivity. Example public prompts can be found in the original repository.

\paragraph{Typical usage}

We assume the user generates content using a generator model – usually an LLM. Inference is then performed using a mix of candidate models (LLMs and SLMs, different from the generator) for benchmarking purposes. Evaluation is conducted by an independent judge model (usually an LLM) to prevent bias from the generation and inference stages. The last step consists in measuring LLM-human judge alignment, to establish whether the LLM judge can be trusted.

\begin{table}[h]
\caption{Taxonomical Classification of Tone Modifications}
\resizebox{\textwidth}{!}{
\begin{tabular}{ll}
\toprule
\textbf{Tone} & \textbf{Description} \\
\midrule
Emojify & Integrates ideographic elements to enhance textual expression while preserving semantic content \\
\addlinespace
Professional & Elevates register to suit formal business and institutional contexts \\
\addlinespace
Shorten & Condenses content through strategic elimination of redundancies \\
\addlinespace
Witty & Incorporates clever wordplay to enhance engagement and rhetorical appeal \\
\addlinespace
Casual & Reduces formality to achieve conversational approachability \\
\addlinespace
Elaborate & Expands content via detailed exposition and contextual enrichment \\
\addlinespace
Proofread & Optimizes grammatical accuracy and orthographic correctness \\
\addlinespace
Improve & Enhances overall communicative efficacy and stylistic quality \\
\addlinespace
Keypoints & Extracts essential informational elements from given content \\
\bottomrule
\end{tabular}
}
\label{table:tones}
\end{table}

\subsection{Base models}

We restrict our analysis to the Phi and Qwen model families. We repurpose the Qwen nomenclature across all model providers for better readability: \texttt{ModelVersion-Size}. Our hypothetical on-device setup is a 1-4B parameter model with an open license. We thus focus~\footnote{Qwen2.5-3B can not be considered because of a special non-commercial license. Note that it is not performing better than others in the same 3-4B family.} our analysis on Phi3~\cite{phi3}, Phi3.5~\cite{phi3}, Phi4~\cite{phi4}, Qwen2~\cite{qwen2}, and Qwen3~\cite{qwen3}. On the cloud, we compare against Haiku 3~\cite{haiku3}, and Haiku 3.5~\cite{haiku35}.

We use the instruction-tuned model version with the lowest context size (usually 4K tokens) for all models and add a no-thinking/reasoning setup for Qwen3 and Phi4. Qwen2.5-3B does not make it to this table because it has a special non-commercial license, but it is not performing better than others in the same 3-4B family on public benchmarks.

\subsection{Public benchmarks}

We found it useful to use public reasoning benchmarks to select candidate models for evaluation and finetuning. We display public reasoning (Table~\ref{tab:public-benchmarks}) and private WA (Table~\ref{tab:main-results}) benchmarks side-by-side in this paper to witness their level of alignment. We selected the reasoning benchmarks that required the least reasoning, thus most akin to the WA tasks: the Open LLM Leaderboard~\cite{openLLM}, MMLU~\cite{mmlu}, MMLU Pro~\cite{mmlupro}, BBH~\cite{bigBench}, IfEval~\cite{ifEval}, and Oogabooga~\cite{oobabooga}. We excluded the mathematics and code benchmarks as they are too far removed from the task at hand. All benchmarks are scored from 0-100 except Oogabooga: it is scored 0-48, indicating the number of correct answers over 48 secret questions that only the anonymous leaderboard owner knows. The Open LLM Leaderboard was useful as an aggregator but was archived in March 2025.

\subsection{Finetuned models}

SLMs demonstrate strong baseline performance on non-reasoning tasks like WA. To further improve performance, we employed Supervised Fine Tuning (SFT)~\cite{SFT} with Low Rank Adaptation (LoRA)~\cite{LoRA}. LoRA is particularly valuable in this context: by swapping LoRA adapters, an on-device model can quickly switch between specialized tasks while maintaining only one base model in memory, addressing strict disk and RAM constraints.

For finetuning, we generated training data using an LLM to create pairs of original sentences and their rewrites (e.g., a sentence and its more professional version). When necessary, we applied Post Training Quantization (PTQ), ensuring we used the same finetuning data to preserve the learned knowledge.

\subsection{Evaluation}

The evaluation framework remains flexible and customizable to accommodate different user requirements. Below, we detail our specific implementations of the LLM and human judgment components, which informed the example prompts provided in the public library.

\textbf{LLM judge} %
The Writing Assist team established a comprehensive quality assessment framework based on four key aspects: accuracy, completeness, coherence, and conciseness. Each aspect is evaluated using specifically designed prompts and a 3-point grading scale, with clear definitions for each grade level. Through this process, we discovered the importance of including an additional validation step to ensure outputs maintained the intended rewrite format rather than defaulting to conversational responses (see Section~\ref{app:outputs}).

\textbf{Human judge} We first asked a set of colleagues to judge the model and provide a score from 0-3: (0) This is not a rewrite. (1) I can't use this rewrite. (2) I would use this rewrite with minor changes. (3) I can use this rewrite as is.

\label{sec:results}
\section{Results}

Looking at the average results in the last column of Table~\ref{tab:main-results}, we notice that the smallest established cloud models tend to dominate our edge candidate models – 4 bits quantized 0.5-4 billion parameter models\footnote{Unfortunately, there is no reliable estimate of the model sizes for Haiku, to put it in perspective. A reasonable assumption is that it is a multiple of the 4B Phi4-4B model.}. Qwen3 provides valuable insights into the impact of model scaling with similar architectures on Writing Assist tasks (i.e., out of the common public benchmark test sets) by offering different small model sizes. From small to medium and medium to large, Qwen3 shows performance increases of $20.7\%$ and $15.3\%$ respectively. The increase is substantial and lifts Qwen3-4B to a level almost on par with common small cloud proprietary models. A similar increase is noticeable between model versions of the same providers and of the same size. 7 months passed between Phi3 (April 2024) and Phi4 (December 2024). The jump we observe is of $14\%$. In other words, within less than a year we can expect a family of 1-2B parameter models to catch up with a family of 4B parameters. Even more surprising, in the same time, we can also expect a family of 4B parameter models to catch up with a cloud model that is expected to be up to an order of magnitude bigger. We note here again that our benchmarks are device-ready models that are 4 bit quantized, which makes this comparison even more surprising. These observations suggest that writing assistance tasks may reach performance plateaus at lower parameter counts than reasoning tasks.

On a per tone basis, we note that models tend to align on which tones they perform the worst at: Emojify, Shorten, and Professional. Qwen3-4B is the exception, where emojify and witty perform on par or better than Haiku3.5.

Finally, public benchmarks (see Table~\ref{tab:public-benchmarks}) tend to match results found with \texttt{wraval}. Phi4 surpasses Phi3.5 on all public and private benchmarks, and Qwen3-4B dominates other small 4 bits quantized models. Despite the general suspicion of models being trained / finetuned on test sets (a.k.a. contamination)~\cite{trainOnTest}, we conclude that public benchmarks are at least reliable to rank models.

\begin{table}[h]
\centering
\caption{LLM-Judge score (0-100) comparison across Writing Assist tones. Non-cloud models are 4 bits quantized for reasonable on-device latency and memory load. SFT refers to supervised finetuning models with LoRA. A small number indicates that the model was not finetuned for that tone. The average tone is calculated over the tuned tones, if available, else over the base-model tones.}
\label{tab:main-results}
\resizebox{\textwidth}{!}{
\begin{tabular}{ll*{9}{S[table-format=2.1]}S[table-format=2.2]}
\toprule
\rotatebox{90}{\textbf{Family}} & \textbf{Model} & {\rotatebox{90}{\textbf{Emojify}}} & {\rotatebox{90}{\textbf{Shorten}}} & {\rotatebox{90}{\textbf{Professional}}} & {\rotatebox{90}{\textbf{Witty}}} & {\rotatebox{90}{\textbf{Casual}}} & {\rotatebox{90}{\textbf{Elaborate}}} & {\rotatebox{90}{\textbf{Improve}}} & {\rotatebox{90}{\textbf{Keypoints}}} & {\rotatebox{90}{\textbf{Proofread}}} & {\rotatebox{90}{\textbf{Avg. Tone}}} \\
\midrule
\multirow{1}{*}{\rotatebox{90}{1B}} & Qwen3-0.6B & 76.5 & 68.6 & 41.1 & 37.0 & 71.3 & 65.1 & 72.4 & 88.9 & 58.3 & 64.36 \\
\midrule
\multirow{2}{*}{\rotatebox{90}{1-2B}} & Qwen2-1.5B & 31.5 & 46.0 & 77.5 & 53.5 & 58.5 & 74.0 & 83.5 & 91.5 & 92.5 & 67.61 \\
& Qwen3-1.7B & 87.3 & 74.6 & 68.8 & 68.8 & 77.2 & 75.5 & 89.7 & 66.6 & 90.5 & 77.67 \\
\midrule
\multirow{6}{*}{\rotatebox{90}{4B}} & Phi3-4B & 45.5 & 46.5 & 72.5 & 61.0 & 87.5 & 78.5 & 86.0 & 96.0 & 91.5 & 73.89 \\
& Phi3.5-4B & 54.0 & 61.2 & 82.2 & 86.1 & 87.7 & 95.2 & 83.9 & 93.0 & 93.1 & 81.82 \\
& Phi3.5-4B-SFT & 80.5 & 73.8 & {\tiny{82.2}} & {\tiny{86.1}} & {\tiny{87.7}} & {\tiny{95.2}} & {\tiny{83.9}} & {\tiny{93.0}} & {\tiny{93.1}} & 86.17 \\
& Phi4-4B & 71.4 & 77.0 & 76.3 & 85.0 & 93.5 & 96.1 & 89.7 & 99.4 & 91.3 & 86.63 \\
& Phi4-4B-SFT & 77.3 & 94.3 & 82.2 & {\tiny{85.0}} & {\tiny{93.5}} & {\tiny{96.1}} & {\tiny{89.7}} & {\tiny{99.4}} & {\tiny{91.3}} & 89.87 \\
& Qwen3-4B & \textbf{96.5} & 61.7 & 86.4 & \textbf{88.9} & 94.2 & 93.1 & 91.3 & 97.7 & 96.4 & 89.58 \\
\midrule
\multirow{2}{*}{\rotatebox{90}{Cloud}} & Haiku3 & 90.0 & 86.0 & \textbf{94.5} & 69.0 & 79.0 & 90.0 & 92.5 & 86.5 & \textbf{97.5} & 87.22 \\
& Haiku3.5 & \textbf{96.5} & \textbf{97.9} & 93.4 & 84.1 & \textbf{98.9} & \textbf{97.1} & 88.7 & \textbf{99.6} & 94.0 & \textbf{94.47} \\
\bottomrule
\end{tabular}
}
\end{table}

\label{sec:learnings}
\section{Learnings}

Our empirical investigation revealed several critical methodological components for efficient model adaptation:

\textbf{Feasibility} It is already possible today to deploy an open source LoRA instruction finetuned and quantized model on devices with a reasonable latency, accuracy and device performance pressure. Hotswapping LoRA adapters for different tasks add to the latency and memory pressure guarantees.

\textbf{LLM Judge Scrutiny} The scrutiny level of LLM judges varies considerably across models, even when using identical prompts. For instance, Sonnet 3.5 would penalize emojified sentences for emoji placement spacing and challenge rephrasing choices, such as treating ``good morning'' and ``Hey'' as semantically different. In contrast, Sonnet 3 demonstrated more lenient evaluation patterns that better aligned with our product and quality assessment teams' standards. Overall, we observed high human-LLM judge alignment, with the notable exception of conversational response detection.

\textbf{A dedicated rewrite judge} was necessary to detect chats that became conversations, instead of rewrites (see examples in Section~\ref{app:outputs}).

\textbf{The importance of humans} can not be overstated~\cite{humansInEval}: starting from the previous remark on the rewrite judge, up to determining what is the desired behavior in edge cases.

\textbf{Synthetic data generation} We observed that LMs are reasonably good at generating heterogeneous data in a structured format with a single prompt, although they tend to generate similar data across models. The single prompt technique, while more economical, does result consistently in the wrong number of examples though: when asked for 100 examples, an LM usually provided 80-90 examples.

\textbf{Finetuning} was particularly useful to nudge the LLM to follow instructions, as opposed to prompt engineering, which has its limitations. Finetuning with just around 100 examples provided up to 10 percentage points improvements. Due to our chipmaker dependencies, we had to finetune on the full precision model before quantizing. 

\section{Outlook}

In the future we would like to refine the judge forms and use prompt optimizers, and compare supervised finetuning with recent reinforcement learning methods. More specifically, we elaborate on four topics below.

\textbf{Multilingual support} We propose a systematic evaluation framework consisting of progressive model testing phases, beginning with proprietary Bedrock cloud-based models, followed by open-source edge candidates, and concluding with specialized multilingual architectures. The evaluation methodology employs both automated LLM-as-judge assessments and human evaluation protocols across multiple writing tones and styles. Supervised fine-tuning approaches could be considered, though preliminary research suggests this path may yield limited improvements~\cite{finetuneMultilingual}.

\textbf{Reinforcement Learning} (RL) is particularly useful when a lot of possible correct ways exist to complete the next sentence. This is true for Writing Assist too. RL would also allow for a more diverse set of behaviors displayed by the Writing Assistant. However, reinforcement learning has recently been shown to be under-performing simpler prompt tuning~\cite{promptOverRL}. Recent publications propose combining SFT and prompt tuning~\cite{SFTPromptTuning}, or RL and prompt tuning~\cite{GRPOPromptTuning}.

Regarding \textbf{prompt tuning}, we intend to integrate with libraries like LMOps~\cite{LMOps} or DSPy~\cite{dspy}. This way \texttt{wraval} can remain prompt-agnostic while helping the user to objectively improve their own prompts. This project also might include a more elaborate prompt versioning system, over the current \texttt{git} solution.

\textbf{More models} In a future publication, we would release numbers on further models to witness differences of architecture scaling within and across model families.

\section{Conclusion}

In this paper, we presented a Writing Assist evaluation framework that uses a specialized judge template designed to assess text modification tasks through an AI assistant specializing in scoring specific text transformations. The evaluation process involved analyzing pairs of original and modified texts along with a specified task (such as "emojify," "professional," or "proofread"), using a detailed scoring rubric. The assessment followed a structured approach where the judge cross-referenced the original and modified texts in relation to the requested tone, provided detailed reasoning, and assigned scores on a 1-3 scale (with 1 being worst and 3 being best). The results were systematically recorded with detailed observations about the output text, focusing on strict adherence to the scoring rubric rather than general evaluation.

\textbf{Ethical Considerations} The deployment of writing assistance models, even small ones, raises important ethical considerations. Text transformation systems may perpetuate biases present in their training data, potentially reinforcing stereotypical language patterns or social biases when modifying text for different tones. For example, "professional" tone transformations might favor Western corporate communication styles, disadvantaging users from different cultural backgrounds. Additionally, these systems could be misused to generate misleading content or manipulate text in ways that obscure its original meaning or intent. Privacy concerns also arise as writing assistance tools process potentially sensitive user content. Our framework supports responsible development by enabling thorough evaluation across diverse writing styles and contexts, helping identify and mitigate these risks before deployment. Future work should explicitly incorporate fairness and bias metrics into the evaluation process.

We hope that this endeavor is helpful for practitioners to quickly test ideas, or to include it in a full product development cycle.

\newpage

\bibliography{iclr2026_conference}
\bibliographystyle{iclr2026_conference}

\appendix

\section{Appendix – Public Benchmarks}

\begin{table}[h]
\centering
\label{tab:public-benchmarks}
\caption{Comparison of Small Language Models across various public reasoning benchmarks}
\begin{tabular}{llccccccccc}
\toprule
Family & Model & \rotatebox{90}{\parbox{2.2cm}{Params (B)}} & \rotatebox{90}{\parbox{2.2cm}{Open LLM}} & \rotatebox{90}{\parbox{2.2cm}{MMLU}} & \rotatebox{90}{\parbox{2.2cm}{MMLU Pro}} & \rotatebox{90}{\parbox{2.2cm}{BBH}} & \rotatebox{90}{\parbox{2.2cm}{IfEval}} & \rotatebox{90}{\parbox{2.2cm}{Oogabooga}} \\
\midrule
\multirow{6}{*}{1-2B}
& \texttt{MobiLlama-1B} & 1.0 & -- & 24.2 & -- & -- & -- & -- \\
& \texttt{Hymba-1.5B} & 1.5 & 13.7 & 52.8 & -- & 4.6 & 57.1 & -- \\
& \texttt{Qwen2-1.5B} & 1.5 & 14.0 & -- & 16.7 & 13.7 & 33.7 & 9.0 \\
& \texttt{SmolLM2-1.7B} & 1.7 & 14.8 & -- & 19.3 & 10.9 & 53.7 & -- \\
& \texttt{Qwen2.5-1.5B} & 1.5 & 15.0 & -- & 20.0 & 19.8 & 44.8 & -- \\
& \texttt{Qwen3-1.7B} & 1.7 & -- & 62.6 & 36.8 & 54.5 & 72.5 & -- \\
\midrule
\multirow{6}{*}{3-4B}
& \texttt{Granite3.1-2B} & 2.5 & 21.1 & -- & -- & -- & -- & -- \\
& \texttt{GeminiNano2} & 3.3 & -- & 55.8 & -- & -- & -- & -- \\
& \texttt{Phi3-4B} & 3.8 & 27.4 & -- & 33.6 & 36.6 & 54.8 & 26.0 \\
& \texttt{Phi3.5-4B} & 3.8 & 27.6 & -- & 32.9 & 36.8 & 57.8 & 18.0 \\
& \texttt{Phi4-4B} & 3.8 & 29.4 & 67.3 & 32.6 & 38.7 & 73.8 & -- \\
& \texttt{Qwen3-4B} & 4.0 & -- & 73.0 & 69.6 & 72.6 & 83.4 & 24 \\
\midrule
\multirow{6}{*}{6.5-8B}
& \texttt{Phi3-7B} & 7.0 & 29.7 & -- & 39.0 & 46.2 & 65.0 & 26.0 \\
& \texttt{Qwen2.5-7B} & 7.0 & 26.9 & 74.2 & 36.5 & 34.9 & 75.9 & 30.0 \\
& \texttt{OLMo2-7B} & 7.0 & 13.2 & 63.9 & 22.7 & 48.5 & 75.6 & -- \\
& \texttt{neo-7B} & 7.0 & -- & 58.3 & 28.7 & -- & -- & -- \\
& \texttt{moxin-7B} & 7.0 & -- & 61.0 & -- & -- & -- & -- \\
& \texttt{Qwen3-8B} & 4.0 & -- & 76.9 & 56.7 & 78.4 & 83 & 24 \\
\midrule
\multirow{2}{*}{Cloud}
& \texttt{Haiku3} & -- & -- & 75.2 & -- & 73.7 & -- & -- \\
& \texttt{Haiku3.5} & -- & -- & 80.3 & 62.1 & 86.6 & 85.9 & -- \\
\bottomrule
\end{tabular}
\end{table}

\section{Appendix – Prompts}

\subsection{Emojify}

Below is an example emojify prompt in, using Qwen conversation tokens.

\begin{lstlisting}
<|im\_start|>system
Emojify this text. Do not convert everything to emojis.<|im\_end|>
<|im\_start|>user
Packing my bags for an exciting trip to the beach! Can't wait to soak up the sun and splash in the waves!<|im\_end|>
<|im\_start|>assistant
Packing my bags for an exciting trip to the beach! [plane] Can't wait to soak up the sun [sun] and splash in the waves! [wave] <|im\_end|>
<|im\_start|>user
The vibrant flowers bloom in spring.<|im\_end|>
<|im\_start|>assistant
\end{lstlisting}

\label{app:outputs}
\section{Appendix – Example Outputs}

Below are example outputs when running 

\begin{lstlisting}
wraval show-results -n 2
\end{lstlisting}

The judge score is shown before the rewrite judge, to illustrate the necessity of an additional judge to detect conversations, instead of rewrites.

\begin{lstlisting}
========================================================================
Tone: casual | Model: qwen-3-1-7B
========================================================================

Example 6418:
Original: I would like to make a reservation for a table near the window.
Rewrite:  Hey, sounds like you're all set for the evening! Let me know if you need help with any other details, like the number of people or any specific requests. I'll be happy to assist! [smile]
Score:    1.50
----------------------------------------

Example 6499:
Original: Could you please turn on the fan?
Rewrite:  No worries, I'll turn it on for you! [thumbsup] Let me know if you need anything else.
Score:    3.00
----------------------------------------

========================================================================
Tone: professional | Model: qwen-3-1-7B
========================================================================

Example 6550:
Original: I was feelin' myself in that outfit, bruh, no lie.
Rewrite:  I was feeling self-assured in that outfit, man, no lie.
Score:    2.50
----------------------------------------

Example 6520:
Original: That movie was a whole vibe, fam.
Rewrite:  I appreciate your feedback, and I'm always eager to hear different perspectives. If you'd like, we could discuss it further!
Score:    1.75
----------------------------------------
\end{lstlisting}

\section{LLM use in paper writing}

First drafts of paragraphs were always written by humans. LLMs (Anthropic Claude Sonnet 4~\cite{claude}) were sometimes used to polish writing, with formulations such as "check for grammatical mistakes", "what would a reviewer at the ICLR conference comment on this?". A human pass and subsequent changes was always necessary.

Perplexity~\cite{perplexity} was used to find related work.

\end{document}

%% file: math_commands.tex
\usepackage{amsmath,amsfonts,bm}

\def\eqref#1{equation~\ref{#1}}

\def\1{\bm{1}}

\DeclareMathAlphabet{\mathsfit}{\encodingdefault}{\sfdefault}{m}{sl}
\SetMathAlphabet{\mathsfit}{bold}{\encodingdefault}{\sfdefault}{bx}{n}